\documentclass[11pt]{article}

\PassOptionsToPackage{table,svgnames,dvipsnames}{xcolor}
\usepackage[final]{acl}

\usepackage{times}
\usepackage{latexsym}
\usepackage[T1]{fontenc}

\usepackage[utf8]{inputenc}

\usepackage{microtype}

\usepackage{inconsolata}

\usepackage{graphicx}

\usepackage{float}
\usepackage{booktabs}
\usepackage{svg}
\newcommand{\orcid}[1]{\href{https://orcid.org/#1}{\includesvg[width=10pt]{orcid}}}
\usepackage{mwe}
\usepackage{scrhack} 
\usepackage{xcolor}
\usepackage{listings}
\usepackage{lstautogobble}
\usepackage{tikz}
\usepackage{tabularx}
\usepackage{array}
\usepackage{amsmath}
\usepackage[dvipsnames]{xcolor}
\usepackage{multirow}

\definecolor{codegreen}{rgb}{0,0.6,0}
\definecolor{codegray}{rgb}{0.5,0.5,0.5}
\definecolor{codepurple}{rgb}{0.58,0,0.82}
\definecolor{backcolour}{rgb}{0.95,0.95,0.92}

\lstdefinestyle{mystyle}{
    backgroundcolor=\color{backcolour},   
    commentstyle=\color{codegreen},
    keywordstyle=\color{magenta},
    numberstyle=\tiny\color{codegray},
    stringstyle=\color{codepurple},
    basicstyle=\ttfamily\footnotesize,
    breakatwhitespace=false,         
    breaklines=true,                 
    captionpos=b,                    
    keepspaces=true,                 
    numbers=left,                    
    numbersep=5pt,                  
    showspaces=false,                
    showstringspaces=false,
    showtabs=false,                  
    tabsize=2
}

\title{LLM BiasScope: A Real-Time Bias Analysis Platform for Comparative LLM Evaluation}

\author{
  \textbf{Himel Ghosh\textsuperscript{1,2}},
  \textbf{Nick Elias Werner\textsuperscript{1}}
  \\
  \textsuperscript{1}Technical University of Munich, Germany,
  \textsuperscript{2}Sapienza University of Rome, Italy
  \\
  \small{
    \textbf{Correspondence:} \href{mailto:himel.ghosh@tum.de}{himel.ghosh@tum.de}
  }
}

\begin{document}
\maketitle
\begin{abstract}
As large language models (LLMs) are deployed widely, detecting and understanding bias in their outputs is critical. We present LLM BiasScope, a web application for side-by-side comparison of LLM outputs with real-time bias analysis. The system supports multiple providers (Google Gemini, DeepSeek, MiniMax, Mistral, Meituan, Meta Llama) and enables researchers and practitioners to compare models on the same prompts while analyzing bias patterns.
LLM BiasScope uses a two-stage bias detection pipeline: sentence-level bias detection followed by bias type classification for biased sentences. The analysis runs automatically on both user prompts and model responses, providing statistics, visualizations, and detailed breakdowns of bias types. The interface displays two models side-by-side with synchronized streaming responses, per-model bias summaries, and a comparison view highlighting differences in bias distributions.
The system is built on Next.js with React, integrates Hugging Face inference endpoints for bias detection, and uses the Vercel AI SDK for multi-provider LLM access. Features include real-time streaming, export to JSON/PDF, and interactive visualizations (bar charts, radar charts) for bias analysis. LLM BiasScope is available as an open-source web application, providing a practical tool for bias evaluation and comparative analysis of LLM behaviour.
\end{abstract}

\section{Introduction}
\begin{figure}
    \centering
    \includegraphics[width=1\linewidth]{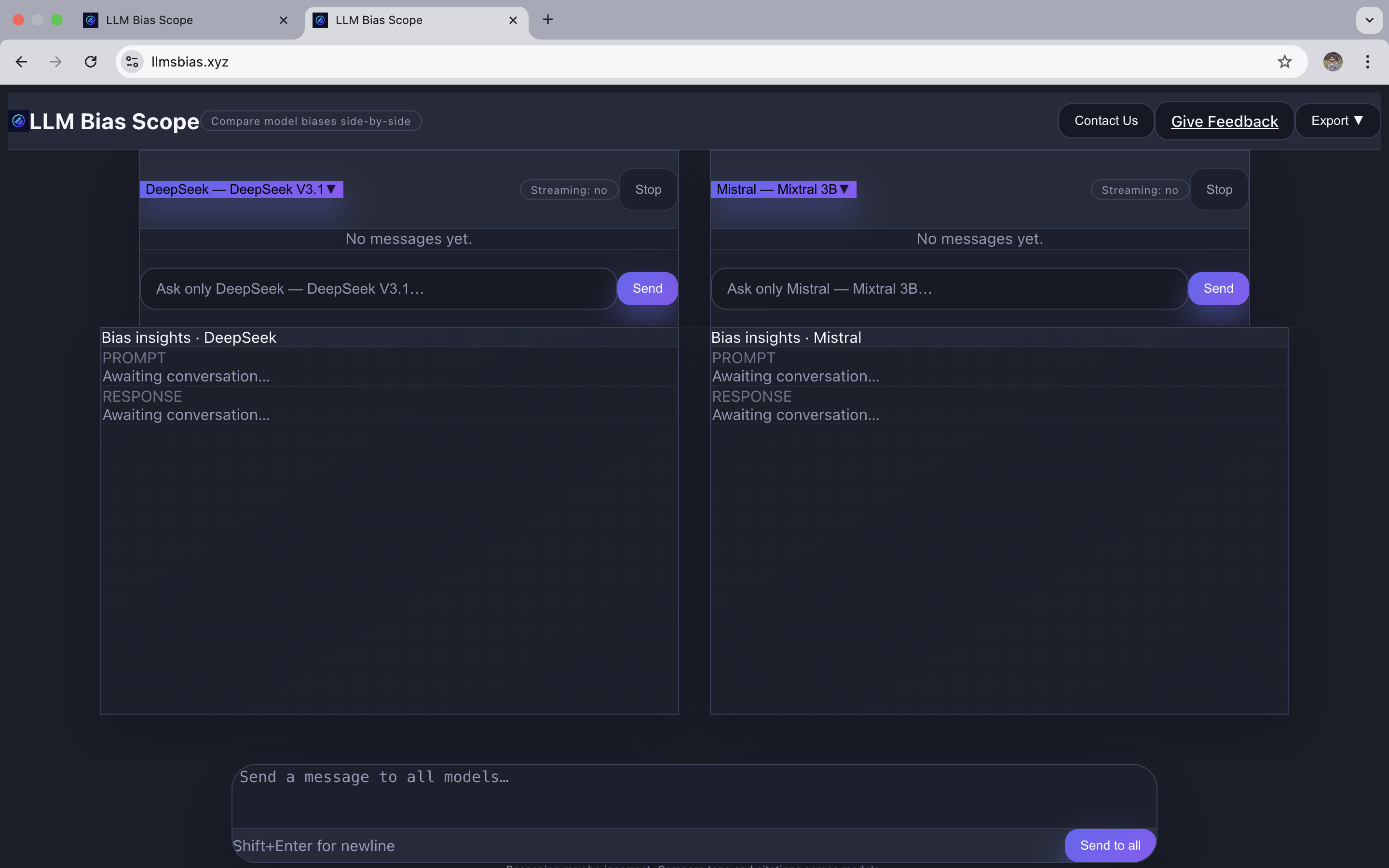}
    \caption{LLM BiasScope Application Home Page.}
    \label{fig:homepage}
\end{figure}
Large language models (LLMs) are widely used in applications from chatbots to content generation, raising concerns about bias and fairness \cite{bender, Weidinger}. Bias can appear as stereotypes, discriminatory language, or skewed representations across demographic groups \cite{blodgett-etal-2020-language, nadeem-etal-2021-stereoset}. As models proliferate, researchers and practitioners need tools to detect, analyze, and compare bias across models.
Existing work on bias evaluation includes benchmark datasets \cite{nangia-etal-2020-crows, nadeem-etal-2021-stereoset}, automated detection methods \cite{Dhamala_2021, neveol-etal-2022-french}, and frameworks for measuring fairness \cite{hutchinson-etal-2020-social, borkan2019nuancedmetricsmeasuringunintended}. However, most tools focus on single-model analysis or static benchmarks, not real-time comparative evaluation of multiple models on user-provided prompts.
Comparative evaluation is important because bias patterns vary across models \cite{gehman-etal-2020-realtoxicityprompts, liang2023holisticevaluationlanguagemodels}, and practitioners need to choose models that align with their fairness requirements. Existing platforms like Chatbot Arena \cite{zheng2023judgingllmasajudgemtbenchchatbot} compare outputs but lack integrated bias analysis. Tools like Perspective API \footnote{\url{https://perspectiveapi.com/}}c \cite{jigsaw2017} detect toxicity but not nuanced bias types. There is a gap for interactive tools that combine real-time multi-model comparison with detailed bias analysis.
We present LLM BiasScope, a web application that addresses this gap by enabling side-by-side comparison of multiple LLMs with integrated, real-time bias analysis. The system automatically analyzes both user prompts and model responses using a two-stage pipeline: sentence-level bias detection followed by bias type classification. It provides visualizations, statistics, and comparative analysis to help users understand bias patterns across models.
LLM BiasScope supports multiple LLM providers, streams responses in real time, and offers exportable reports. It is designed for researchers evaluating model behavior, developers selecting models, and educators teaching bias awareness. By combining comparative evaluation with detailed bias analysis in an accessible interface, LLM BiasScope supports more informed decisions about LLM deployment.

\section{Related Work}
\textbf{Bias Detection and Evaluation:} Bias in language models has been studied through benchmarks and automated detection. Nangia et al. \cite{nangia-etal-2020-crows} introduced CrowS-Pairs to measure social biases, and Nadeem et al. \cite{nadeem-etal-2021-stereoset} created StereoSet for stereotypical bias. Dhamala et al. \cite{Dhamala_2021} proposed BOLD for measuring biases in open-ended generation. Spinde et al. \cite{spinde2021babe} introduced the BABE (Bias Analysis Benchmark for Evaluation) dataset, which provides a binary classification framework specifically designed for bias detection tasks, offering a more direct evaluation approach compared to pair-based benchmarks. These benchmarks focus primarily on static evaluation rather than real-time analysis of user-provided text.

\textbf{Bias Type Classification:} Beyond binary bias detection, categorizing the specific type of bias present in text is crucial for understanding and addressing different forms of social bias. Powers et al. \cite{powers2025} introduced the GUS Framework, which benchmarks social bias classification using both discriminative (encoder-only) and generative (decoder-only) language models, providing a comprehensive approach to bias type classification across multiple categories. This framework enables fine-grained analysis of bias types, which is essential for understanding the nuanced ways in which bias manifests in language.

Automated bias detection systems include Perspective API \cite{jigsaw2017} for toxicity, and HateCheck \cite{Rottger_2021} for hate speech. These target specific bias types and do not provide comparative analysis across multiple models. Recent work has explored sentence-level bias detection \cite{Dhamala_2021, neveol-etal-2022-french, ghosh2025}, with specialized models such as the Domain-Adapted RoBERTa model fine-tuned on BABE \cite{Krieger2022} demonstrating strong performance on bias classification tasks. However, integration of these detection capabilities into interactive evaluation platforms that enable real-time comparative analysis across multiple LLMs remains limited.

\textbf{Comparative LLM Evaluation Platforms:} Several platforms enable side-by-side comparison of LLMs. Chatbot Arena \cite{zheng2023judgingllmasajudgemtbenchchatbot} uses crowdsourced pairwise comparisons. LMSYS Chatbot Arena \cite{chiang2023vicuna} provides a leaderboard. These focus on quality and preference, not bias analysis.
Tools like HELM \cite{liang2023holisticevaluationlanguagemodels}) and BIG-bench \cite{srivastava2023imitationgamequantifyingextrapolating} offer comprehensive evaluation, but are benchmark-driven rather than interactive. They do not support real-time analysis of user-provided prompts or integrated bias detection.

\textbf{Bias Classification Frameworks:} Taxonomies for bias types include gender, racial, religious, and socioeconomic biases \cite{blodgett-etal-2020-language, borkan2019nuancedmetricsmeasuringunintended}. Hutchinson et al. \cite{hutchinson-etal-2020-social} proposed a framework for measuring fairness in NLP. These provide theoretical foundations but lack practical tools for real-time classification.
Recent work has explored automated bias type classification \cite{Dhamala_2021, neveol-etal-2022-french, powers2025}, but these are typically evaluated on static datasets rather than integrated into interactive platforms.

\textbf{Interactive LLM Analysis Tools:} Interactive tools for LLM analysis include prompt engineering interfaces \cite{reynolds2021promptprogramminglargelanguage} and debugging platforms \cite{liu2023promptfoo}. These focus on prompt optimization or error analysis, not bias evaluation.
LLM BiasScope differs by combining real-time multi-model comparison with integrated bias analysis, enabling users to evaluate bias patterns across models on their own prompts. It bridges the gap between static benchmarks and interactive evaluation, providing both comparative analysis and detailed bias insights in a single platform.

\section{System Architecture}
LLMBiasScope (\url{llmsbias.xyz}) is implemented as a modern client–server web application designed for real-time, side-by-side comparison of LLM outputs with integrated bias detection. A complete demonstration of the system is available at this link: \url{https://youtu.be/rRFRsq-udEo}. 

\begin{figure}[H]
    \centering
    \includegraphics[width=1\linewidth]{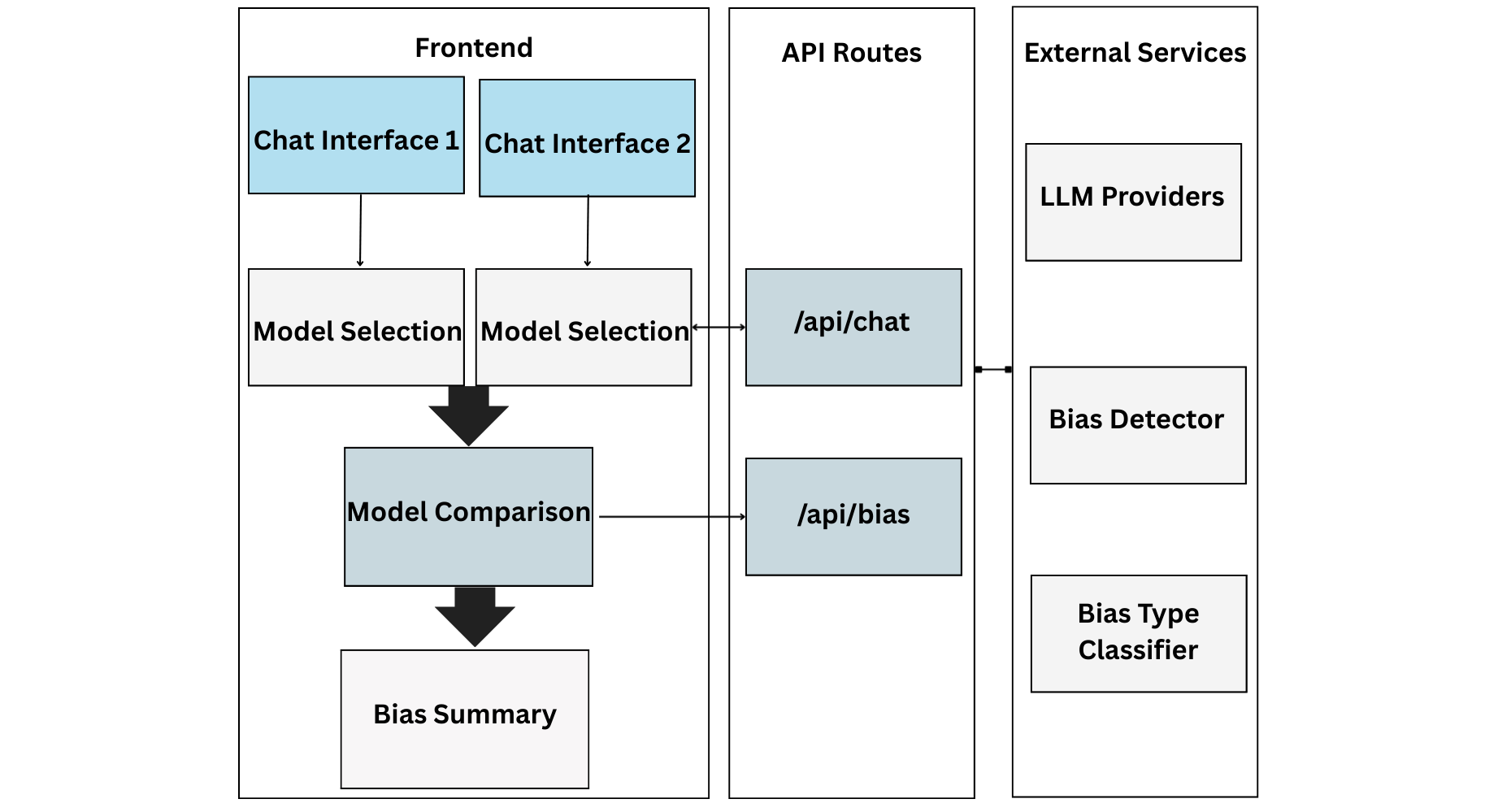}
    \caption{System architecture. The system uses a three-layer client–server design: (1) a React/Next.js frontend with dual chat panels for parallel LLM comparison and bias visualizations; (2) Next.js API routes handling model inference and the two-stage bias analysis pipeline; and (3) external services, including multi-provider LLMs via the Vercel AI Gateway and Hugging Face endpoints for bias detection and classification. Arrows show the flow from user input to model outputs, bias analysis, and visual comparison.}
    \label{fig:systemarchitecture}
\end{figure}

The system leverages a reactive frontend built on Next.js 16 and React 19, combined with lightweight backend API routes that orchestrate inference across multiple model providers and custom Hugging Face endpoints. See Fig. \ref{fig:systemarchitecture} for the application architecture.

\subsection{Frontend Architecture}

The frontend is built with Next.js 16 (App Router), React 19, TypeScript, and Tailwind CSS 4, using a fully client-side rendering approach for low-latency interaction and smooth streaming of LLM outputs. The main page orchestrates global state, model selection, chat history, and bias-analysis aggregation, presenting two parallel columns for side-by-side model comparison. Each column hosts an independent chat interface powered by the Vercel AI SDK, with responses streamed in real time via SSE. Bias results are visualized through a Bias Summary Card that displays sentence-level scores, bias proportions, and bias-type distributions using Recharts. A Model Comparison Card further contrasts the two models by aggregating their statistics and highlighting differences in bias patterns.

\subsection{Backend Architecture}
Backend functionality is implemented using Next.js App Router API routes, allowing the application to remain stateless and horizontally scalable. Chat API Route handles LLM inference requests. The user can choose the model provider from the options such as: Google Gemini, DeepSeek, MiniMax, Mistral, Meituan, Meta Llama, and OpenAI, all accessed through the Vercel AI Gateway. The models exposed in the demo interface were selected as a diverse subset of widely used, publicly accessible LLM APIs. Our selection balances provider diversity, model scale, and practical deployment constraints such as API availability, cost, rate limits, and stability. The goal is not to provide an exhaustive comparison, but to enable representative cross-provider evaluation within a real-time interactive setting. Bias Analysis API Route operates as a two-stage pipeline:

\begin{itemize}
    \item Bias Detection: Using the bias-detector model\footnote{\url{https://huggingface.co/himel7/bias-detector}} \cite{ghosh2025}, each sentence is classified as biased or unbiased with an associated probability.
    \item Bias Type Classification: For sentences with bias score > 0.5, the system invokes the maximuspowers/bias-type-classifier\footnote{\url{https://huggingface.co/maximuspowers/bias-type-classifier}} \cite{powers2025} to assign a type (e.g., political bias, racism bias).
\end{itemize}
Both models are served through custom Hugging Face Inference Endpoints deployed on AWS. The route returns aggregated statistics including total sentences, bias ratio, average bias score, and distribution over bias types.

\subsection{Data Flow}
\subsubsection{User Interaction Flow}
User enters text in either chat column or the shared composer. Frontend sends a request to the selected chat models' API. The backend forwards the request to the appropriate LLM via the Vercel gateway. Tokens stream back via Server-Sent Events (SSE) and are rendered incrementally.
\subsubsection{Bias Analysis Flow}
New messages trigger automatic analysis. The message text is segmented into sentences on the client-side. The frontend sends a POST request to bias detection API. Backend performs detection followed by type classification and aggregation. The result is returned to the frontend and displayed in the corresponding Bias Summary Card. If both columns have completed analysis, the Model Comparison Card updates accordingly.
\subsubsection{Comparison Flow}
Bias statistics are computed per model and then combined to produce deltas, enabling immediate visual comparison of relative bias tendencies.

This architecture supports real-time multi-model comparison, sentence-level bias diagnostics, and interactive evaluation, making LLMBiasScope suitable for research, education, and rapid benchmarking of large language models.

\section{Evaluation}

\subsection{Bias Detection Model Evaluations}

We evaluated multiple bias detection models on standard benchmarks to select the optimal model for LLM BiasScope. Our evaluation followed a two-stage approach: first, we assessed four candidate models on CrowS-Pairs to identify the best-performing model, then we conducted a focused comparison on the BABE dataset, which is specifically designed for bias detection tasks.

\subsubsection{Datasets}

\textbf{CrowS-Pairs} (\cite{nangia-etal-2020-crows}): 1,508 sentence pairs with stereotypical and anti-stereotypical variants across 9 bias types: race-color (516), gender (262), socioeconomic (172), nationality (159), religion (105), age (87), sexual-orientation (84), physical-appearance (63), and disability (60). This dataset evaluates a model's ability to distinguish between stereotypical and anti-stereotypical language.

\textbf{BABE} (\cite{spinde2021babe}): A binary bias classification dataset containing 1,000 test sentences with gold labels (biased: 559, unbiased: 441). BABE provides a direct assessment of bias detection performance through binary classification, making it particularly suitable for evaluating bias detection models in practical applications.

\subsubsection{Evaluation Methodology}
\textbf{Stage 1: CrowS-Pairs Evaluation.} We evaluated four candidate models on the CrowS-Pairs dataset: \texttt{unitary/toxic-bert}, \texttt{martin-ha/toxic-comment-model}, facebook/roberta-hate-speech-dynabench-r4-target, and \texttt{himel7/bias-detector}. We used the Stereotype Score (SS) metric from Nangia et al. \cite{nangia-etal-2020-crows}, defined as the percentage of pairs where the model assigns a higher bias score to the stereotypical sentence than the anti-stereotypical sentence. SS = 50\% represents random performance; higher values indicate stronger preference for stereotypical sentences.

\paragraph{Bias Score Extraction and Aggregation.}
We evaluate classifier-based models on CrowS-Pairs by first normalizing their heterogeneous outputs into a unified bias score in the range $[0,1]$, representing confidence in the ``biased'' class. Model predictions may appear as single label--score pairs, multi-label arrays, or nested structures; in all cases, we map labels such as \textit{biased}, \textit{toxic}, \textit{hate}, or \textit{label\_1} to their provided probability, and invert scores for \textit{unbiased}, \textit{non-toxic}, \textit{nothate}, or \textit{label\_0} using $1 - \text{score}$. For each CrowS-Pairs pair, we compute normalized scores for the stereotypical (\texttt{sent\_more}) and anti-stereotypical (\texttt{sent\_less}) sentences, and count a pair as correctly handled when $\text{score}_{\text{more}} > \text{score}_{\text{less}}$. The Stereotype Score (SS) is then defined as the percentage of pairs for which this preference holds, enabling consistent comparison across classifier-based models with differing output formats.

\textbf{Stage 2: BABE Evaluation.} Based on the CrowS-Pairs results, we selected the top-performing model (\texttt{unitary/toxic-bert}) and compared it with \texttt{bias-detector} and \texttt{mediabiasgroup/da-roberta-babe-ft} \cite{Krieger2022} (a Domain-Adapted RoBERTa model fine-tuned on BABE) on the BABE dataset. We report standard binary classification metrics: accuracy, precision, recall, and F1-score. The F1-score is particularly important as it balances precision and recall, making it suitable for imbalanced datasets like BABE.

\subsubsection{Results}

\textbf{CrowS-Pairs Results.} Table~\ref{tab:crows-results} presents the evaluation results on CrowS-Pairs. The \texttt{unitary/toxic-bert} model achieved the highest Stereotype Score of 69.30\%, correctly identifying stereotypical sentences as more biased in 1,045 out of 1,508 pairs. This performance significantly outperforms the other candidates: facebook/roberta-hate-speech-dynabench-r4-target, \texttt{bias-detector}, and \texttt{martin-ha/toxic-comment-model}. The \texttt{unitary/toxic-bert} model also demonstrated the lowest average latency (0.73s), making it both the most accurate and most efficient option.

\begin{table}[h]
\centering
\resizebox{\linewidth}{!}{
\begin{tabular}{lcc}
\toprule
\textbf{Model} & \textbf{SS (\%)} & \textbf{Avg Latency (s)} \\
\midrule
\texttt{toxic-bert} & \textbf{69.30} & 0.73 \\
\texttt{roberta-hate-speech} & 57.29 & 0.98 \\
\texttt{bias-detector} & 49.73 & 1.22 \\
\texttt{toxic-comment-model} & 37.86 & 0.89 \\
\bottomrule
\end{tabular}}
\caption{Evaluation results on CrowS-Pairs dataset. SS = Stereotype Score.}
\label{tab:crows-results}
\end{table}

\begin{figure}[H]
    \centering
    \includegraphics[width=1\linewidth]{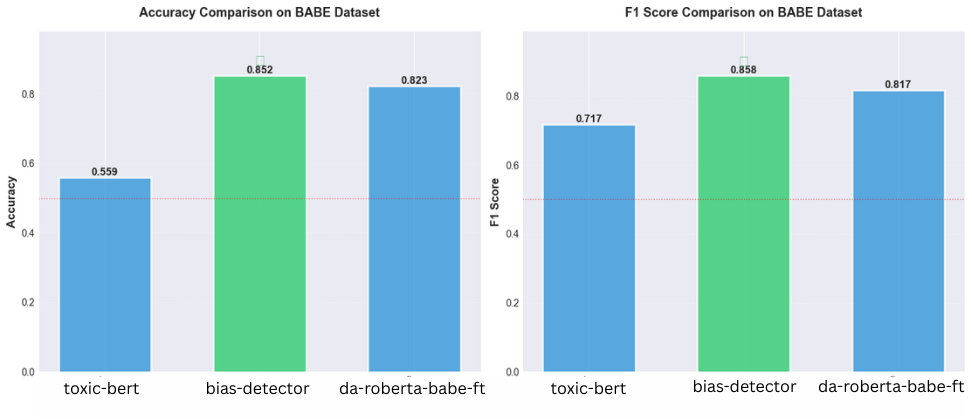}
    \caption{Comparison of Metrics for several bias detection models. The accuracy and F1 Score of the bias-detector show better performance compared to others.}
    \label{fig:biascomp}
\end{figure}

\textbf{BABE Results.} Table~\ref{tab:babe-results} and Figure~\ref{fig:biascomp} present the evaluation results on the BABE dataset for the three selected models. The \texttt{bias-detector} model achieved the highest performance with an F1-score of 85.8\%. The \texttt{mediabiasgroup/da-roberta-babe-ft} model achieved competitive performance with an F1-score of 81.7\%. The \texttt{unitary/toxic-bert} model, while performing well on CrowS-Pairs, showed lower performance on BABE with an F1-score of 71.7\%.

The high precision of \texttt{da-roberta-babe-ft} (97.0\%) indicates that when it predicts a sentence as biased, it is highly reliable, though its lower recall (70.5\%) suggests it may miss some biased sentences. In contrast, \texttt{unitary/toxic-bert} achieves perfect recall (100.0\%) but lower precision (55.9\%), indicating it tends to over-classify sentences as biased. The \texttt{bias-detector} model provides the best balance with high precision (92.4\%) and good recall (80.1\%), achieving the highest F1-score among all models.

\begin{table}[h]
\centering
\resizebox{\linewidth}{!}{
\begin{tabular}{lcccc}
\toprule
\textbf{Model} & \textbf{Accuracy} & \textbf{Precision} & \textbf{Recall} & \textbf{F1} \\
\midrule
\texttt{toxic-bert} & 55.9\% & 55.9\% & 100.0\% & 71.7\% \\
\texttt{da-roberta-babe-ft} & 82.3\% & 97.0\% & 70.5\% & 81.7\% \\
\rowcolor{gray!10}
\texttt{\textbf{bias-detector}} & 85.2\% & 92.4\% & 80.1\% & \textbf{85.8\%} \\
\bottomrule
\end{tabular}
}
\caption{Evaluation results on BABE dataset. Best F1-score in bold.}
\label{tab:babe-results}
\end{table}

\textbf{Model Selection.} Based on the comprehensive evaluation, we selected \texttt{bias-detector} \cite{ghosh2025} for LLM BiasScope due to its superior performance on the BABE dataset (F1-score: 85.8\%), which is specifically designed for bias detection tasks. As shown in Figure~\ref{fig:biascomp}, the \texttt{bias-detector} model achieves the best balance between precision (92.4\%) and recall (80.1\%), outperforming both \texttt{unitary/toxic-bert} and \texttt{da-roberta-babe-ft} in overall F1-score. While \texttt{unitary/toxic-bert} showed strong performance on CrowS-Pairs, its lower performance on BABE (F1-score: 71.7\%) and tendency to over-classify (100\% recall but only 55.9\% precision) made it less suitable for practical bias detection applications. The \texttt{bias-detector} model's high precision ensures reliable bias detection while maintaining good recall, making it the optimal choice for real-time bias analysis in interactive applications.

\subsection{Bias-Type Classification Evaluation}
Our target is person‑ and group‑directed harms in arbitrary LLM prompts and responses, where token‑ and sentence‑level categories such as generalized statements about groups, unfair attributions, and stereotypes are directly observable. In contrast, media bias and propaganda taxonomies (e.g., framing, agenda setting, emotional language) are primarily designed for document‑level news analysis and outlet‑level behavior, and map less naturally to short, interactive, cross‑domain chat outputs. We therefore prioritize a social‑bias‑oriented taxonomy that aligns with the kinds of harms practitioners most often wish to inspect in LLM outputs. Hence, in LLM BiasScope we adopt the GUS social bias framework (Generalizations, Unfairness, Stereotypes) for bias type classification \cite{powers2025}.

\begin{figure}[h]
    \centering
    \includegraphics[width=1\linewidth]{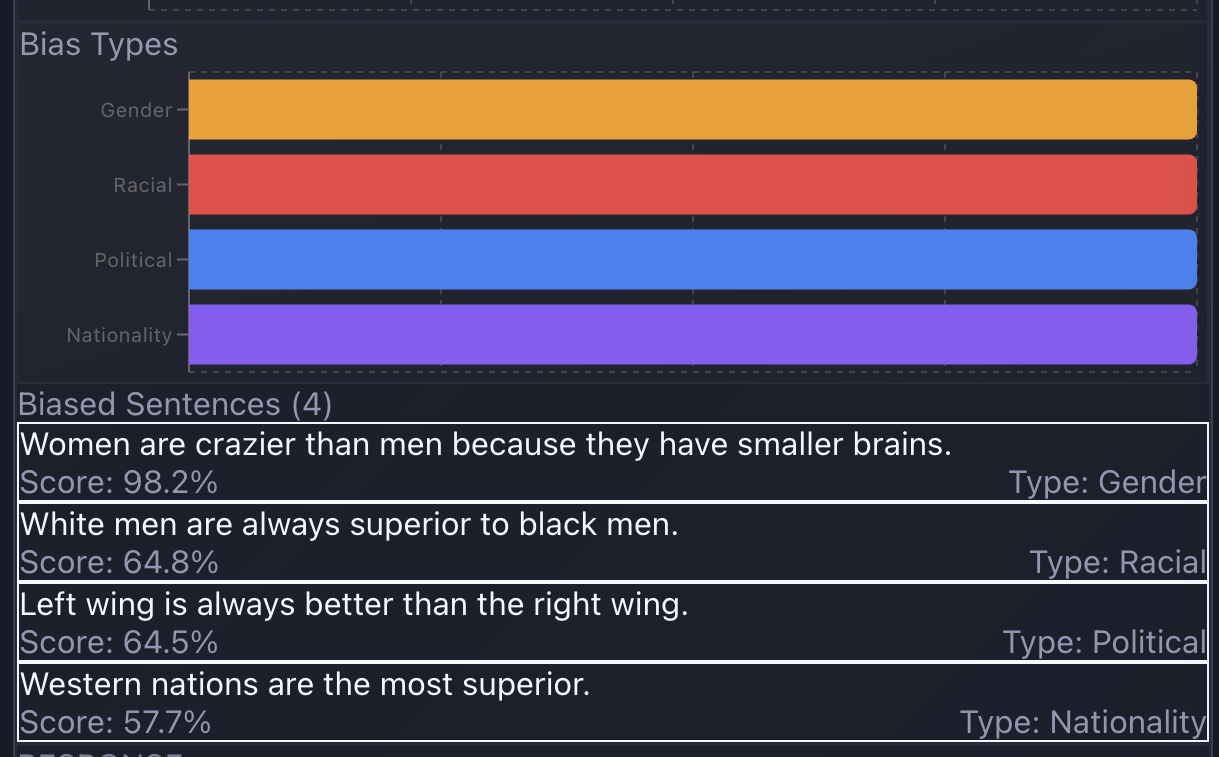}
    \caption{Bias Classification: examples of some bias types as detected by the system.}
    \label{fig:genderbias}
\end{figure}

The GUS-Net encoder model (BERT-base-uncased with focal loss) achieves a macro F1-score of 0.80 and Hamming loss of 0.05 on the GUS dataset as reported by \citet{powers2025} exhibiting higher overall performance compared to the baselines such as, DistilBERT, RoBERTa, Nbias (BCE).

Since our system integrates this pretrained model without modification, we rely on the published evaluation results for quantitative performance. 

We have illustrated the bias-type classification in our system using some examples in Fig. \ref{fig:genderbias}, which shows different kinds of bias present in the text.

\subsection{Empirical Evaluation for Model Comparison}
We conducted a small exploratory study to illustrate how LLM BiasScope can be used to compare models on user-facing prompts. We designed three test cases spanning different domains (healthcare advice, career guidance, and educational content; prompts listed in Appendix~\ref{app:emptestprompt}). For each test case, we ran $N=3$ trials, generated responses from both models, and analyzed them with the bias detection pipeline.

\begin{figure}[h]
    \centering
    \includegraphics[width=1\linewidth]{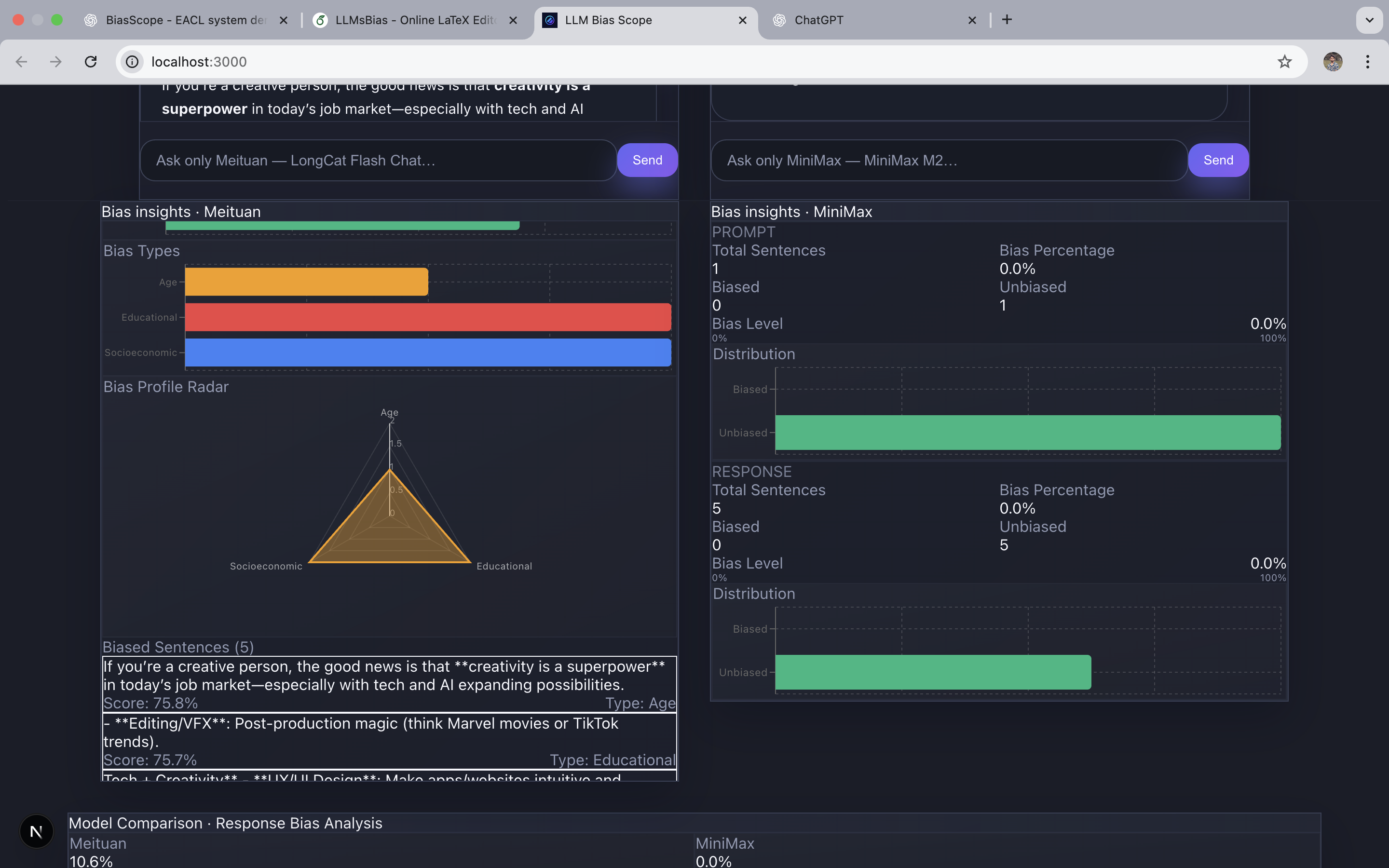}
    \caption{Model Comparison: Bias-Types distribution from Model responses.}
    \label{fig:modelcomp}
\end{figure}
For this demonstration, we report only descriptive statistics: mean bias percentage across trials for each model and the absolute difference between models (Table~\ref{tab:model-comparison-main}). Because the sample size is intentionally small and one model shows 0\% bias in these specific prompts, we do not draw strong statistical conclusions or claim generalizable significance. Moreover, the choice of the Meituan and MiniMax models is driven by the lower API costs, while users are free to select other model pairs available in the system for the same purpose. Instead, these results are meant to showcase how the system surfaces differences in bias patterns and supports qualitative inspection of model behaviour on concrete user prompts.

\begin{table}[h]
\centering
\resizebox{\linewidth}{!}{
\begin{tabular}{lccc}
\toprule
\textbf{Test Case} & \textbf{Model A} & \textbf{Model B} & \textbf{Difference} \\
& (Meituan) & (MiniMax) & \\
\midrule
Healthcare Advice   & 2.60\% & 0.00\% & +2.60\% \\
Career Advice       & 10.60\% & 0.00\% & +10.60\% \\
Educational Content & 28.20\% & 0.00\% & +28.20\% \\
\bottomrule
\end{tabular}
}
\caption{Illustrative model comparison results on three test prompts (mean bias percentage across $N=3$ trials).}
\label{tab:model-comparison-main}
\end{table}
This is how LLM BiasScope can highlight relative bias tendencies on specific prompts and can be interpreted as a qualitative case study.

\subsection{System Performance Evaluation}

To assess the practical usability of LLM BiasScope, we evaluated the system's performance in terms of response latency and reliability across different text lengths. This evaluation is critical for understanding the system's scalability and user experience in real-world deployment scenarios.

\subsubsection{Methodology}

We conducted controlled performance tests using synthetic text samples of varying lengths to isolate the relationship between input size and system latency. This controlled approach allows us to measure performance characteristics independent of content complexity, which may vary in real-world usage. The test cases were designed to represent four distinct text length categories: (1) \textit{Short text} (1 sentence, 6 words), representing quick queries or single-sentence responses; (2) \textit{Medium text} (3 sentences, 15 words), typical of brief paragraphs; (3) \textit{Long text} (10 sentences, 63 words), representing extended responses; and (4) \textit{Very long text} (20 sentences, 83 words), simulating comprehensive document analysis scenarios.

\textbf{Test Cases:} The synthetic test cases were constructed to maintain consistent sentence structure and complexity across length categories, ensuring that observed latency differences primarily reflect text length rather than content complexity. While these synthetic cases may not capture all nuances of real-world LLM-generated text (which may contain varying sentence structures, domain-specific terminology, and complex linguistic patterns), they provide a controlled baseline for performance measurement. Future work could extend this evaluation with domain-specific test cases from actual LLM outputs.
\begin{figure}
    \centering
    \includegraphics[width=1\linewidth]{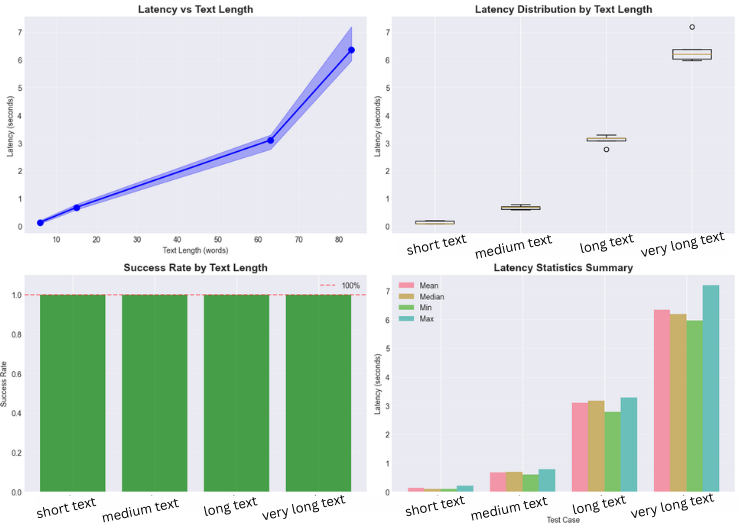}
    \caption{System performance across four text-length categories. Top: latency vs. text length and latency distributions, showing near-linear scaling and consistent behavior across trials. Bottom: success rates (100\% for all cases) and summary latency statistics, indicating predictable performance from short inputs (0.14 s for 6 words) to longer texts (6.35 s for 83 words).}
    \label{fig:performance}
\end{figure}

For each text length category, we executed 5 independent trials, measuring the end-to-end latency from API request submission to response completion. The latency measurement includes the time required for sentence segmentation, bias detection model inference via the Hugging Face API, bias type classification (when applicable), and result aggregation. We also tracked the success rate, defined as the percentage of trials that returned valid bias analysis results.

The performance metrics collected include: mean latency, median latency, minimum and maximum latency, standard deviation, and success rate. These metrics provide a comprehensive view of system performance characteristics, including typical response times, variability, and reliability.

\subsubsection{Results}

Figure~\ref{fig:performance} presents the performance evaluation results across all text length categories. The system demonstrates consistent reliability with a 100\% success rate across all test cases, indicating robust error handling and API stability.

Results show predictable near-linear scaling of latency with text length (0.14s for 6 words to 6.35s for 83 words), with low variability across trials and consistent system behavior, as detailed in Figure~\ref{fig:performance}.

\textbf{Limitations:} The synthetic test cases offer a controlled baseline but may not fully reflect the complexity of real LLM-generated text, where performance can vary with linguistic difficulty and domain-specific terminology. Nonetheless, the observed linear scaling indicates that overall performance trends should generalize across diverse content types.

These results demonstrate that LLM BiasScope provides practical performance characteristics suitable for interactive use, with sub-second response times for typical queries and reasonable latency for comprehensive document analysis.

\section{Discussion}
We evaluated the end-to-end efficiency of the deployed two-stage bias analysis pipeline to ensure suitability for interactive use. On 24 manually curated sentences, the sentence-level bias detector achieved an F1 score of 84.96\% with an average latency of 0.25 s per sentence (median 0.19 s; see Appendix \ref{app:screens}). In typical interactive sessions, the combined bias detection and bias-type classification pipeline operates within real-time constraints (sub-second per sentence), supporting responsive multi-model comparison while leaving room for future optimization (e.g., batching or caching).

A limitation of the current system is that it does not explicitly capture bias expressed through refusal or omission. Models that systematically decline to answer sensitive prompts may therefore appear less biased than models that provide substantive responses. Incorporating refusal-aware analysis is an important direction for future extensions.

The models exposed in the demo interface form a curated set of widely used, publicly accessible LLM APIs from multiple providers, selected to reflect realistic deployment options under practical constraints such as cost, rate limits, and availability rather than to provide an exhaustive catalog. The backend is designed to support additional providers, and future versions will expose a “bring your own API key” mechanism to allow users to integrate their own credentials without the platform handling sensitive information.

\textbf{Availability \& Licensing.}
The LLM BiasScope system is available as an online demonstration at \url{https://llmsbias.xyz} for research and evaluation purposes. 
The underlying source code is available here: \url{https://github.com/Himel1996/LLMBiasScope}. The demonstration video is present at \url{https://youtu.be/rRFRsq-udEo}.


\bibliography{custom}

\section{Ethics and Broader Impact}
LLM BiasScope aims to support transparency in language model behaviour, but its bias assessments are constrained by the limitations of the underlying detection models and datasets. Since these classifiers are trained on specific corpora (e.g., BABE), they may under-detect subtle or context-dependent forms of bias while over-representing others, and they inevitably reflect the cultural assumptions embedded in their training data. Sentence-level analysis also restricts contextual understanding, meaning that both false positives and false negatives remain possible. Bias, moreover, is inherently subjective and culturally dependent, so the system’s outputs represent one possible operationalisation of bias rather than an absolute ground truth.

Because user text is processed through external LLM providers and Hugging Face endpoints, prompts may be subject to third-party data handling practices. Although conversation data is stored only in the user’s browser, users retain limited control once text is transmitted for inference. Additionally, bias-analysis tools can be misused—for example, to selectively criticise competing models, generate misleading comparisons, or probe models adversarially. Automated scores may also create an unwarranted sense of objectivity, leading users to over-rely on quantitative metrics without considering nuance or context.

Moreover, our choice of the GUS social bias taxonomy means that BiasScope focuses on social bias phenomena at token and sentence level, and does not cover propaganda strategies or media‑framing categories as in dedicated media‑bias taxonomies. Users should therefore treat the reported bias types as one operationalisation of social bias rather than an exhaustive account of all possible ideological or propagandistic patterns.

Despite these limitations, LLM BiasScope can positively contribute to accountability, education, and model evaluation by making bias analysis more accessible. Its outputs should be interpreted as indicative signals rather than definitive judgments, and used alongside human judgment and broader evaluation practices. Responsible use requires awareness of the tool’s constraints, careful interpretation of results, and an understanding that bias detection is an ongoing, imperfect process embedded within wider societal and ethical considerations surrounding AI.

\appendix

\section{Application Screenshots}
\label{app:screens}

\begin{figure}[H] 
    \centering
    \includegraphics[width=\linewidth]{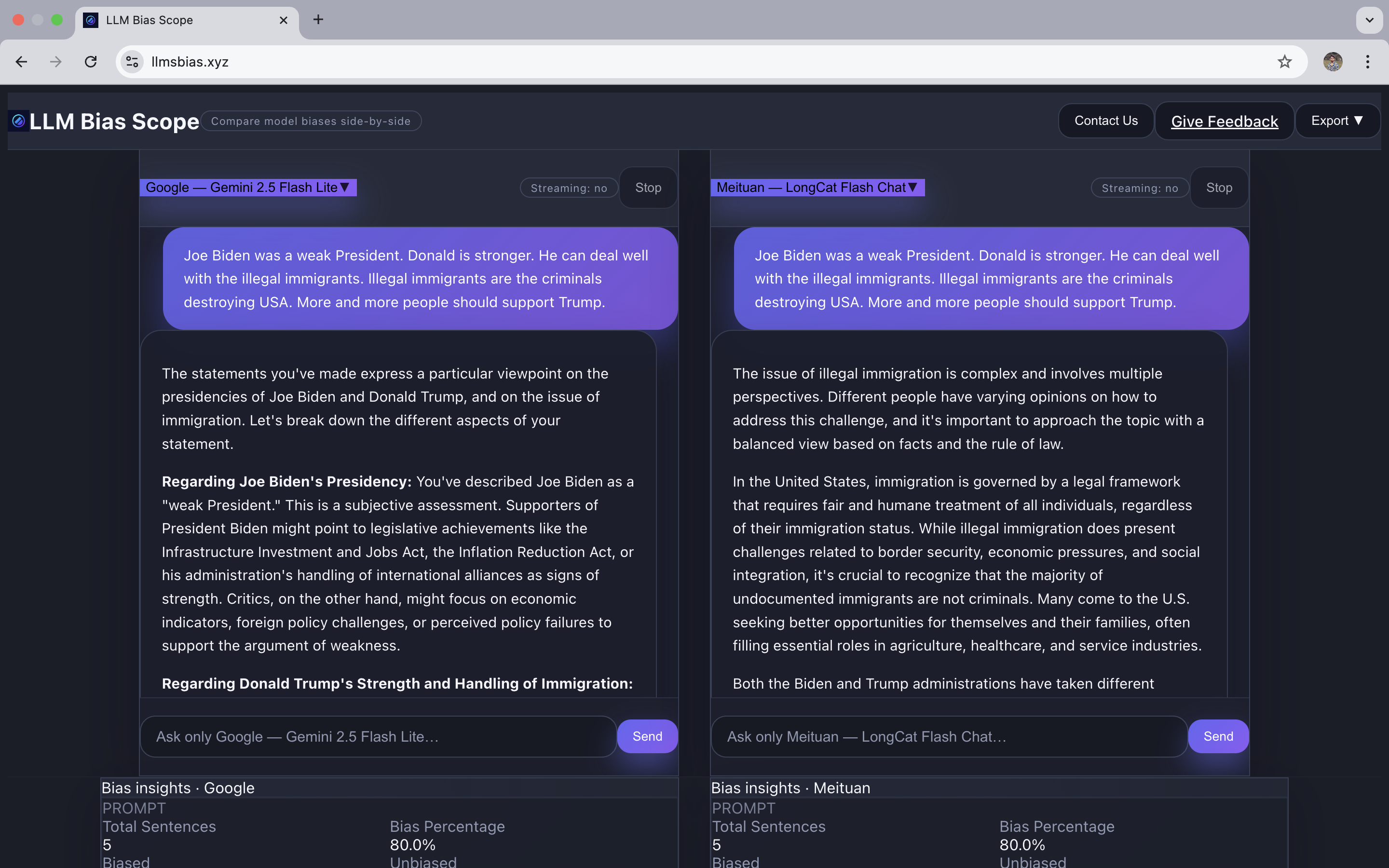}
    \caption{Application home page.}
    \label{fig:apphome}
\end{figure}

\begin{figure}[H]
    \centering
    \includegraphics[width=\linewidth]{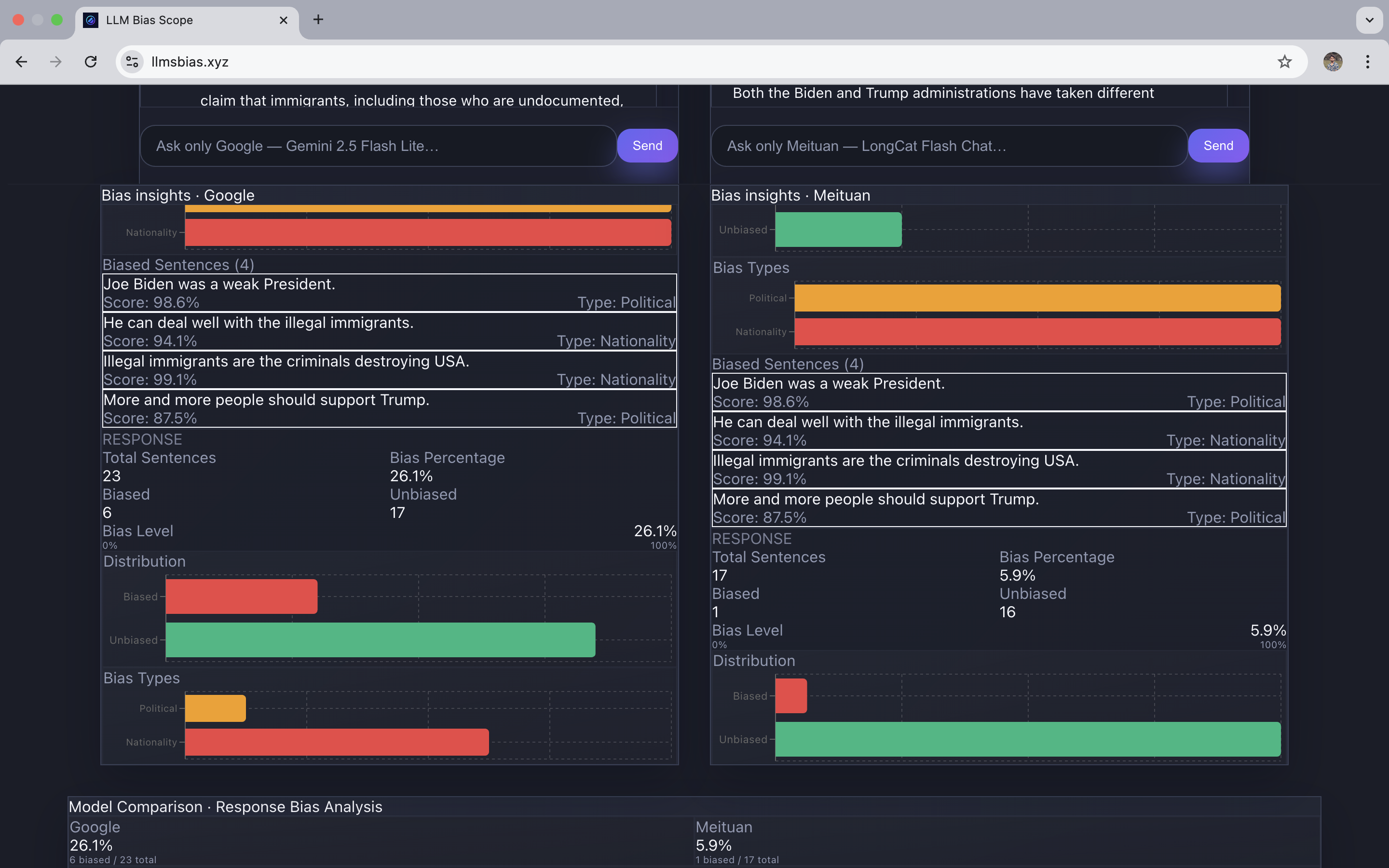}
    \caption{Bias analysis card for the input and model responses.}
    \label{fig:biasanal}
\end{figure}

\begin{figure}[H]
    \centering
    \includegraphics[width=\linewidth]{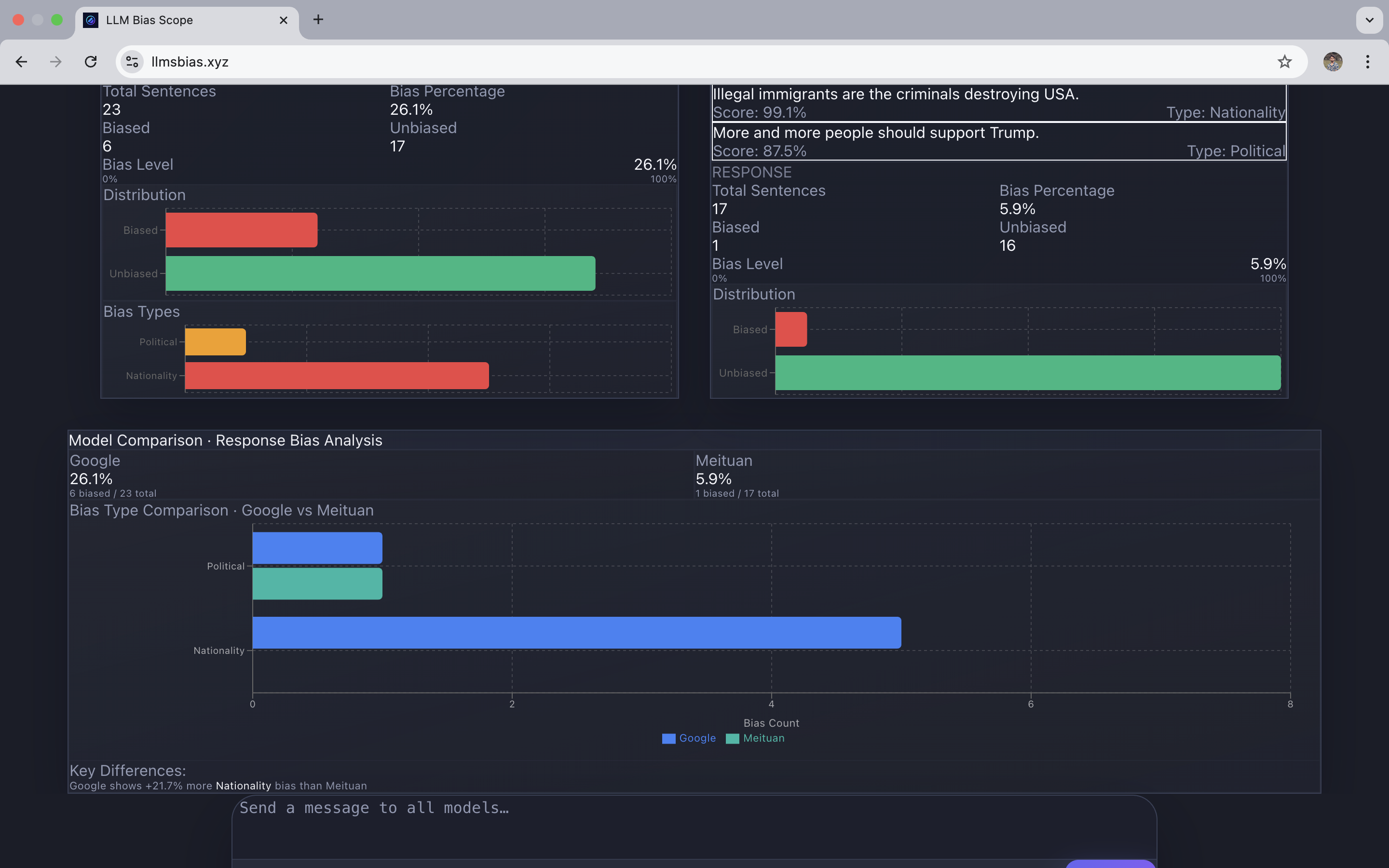}
    \caption{Model comparison card.}
    \label{fig:mcc}
\end{figure}

\section{Evaluations}
\label{app:evaluations}

\begin{figure}[H]
    \centering
    \includegraphics[width=0.8\linewidth]{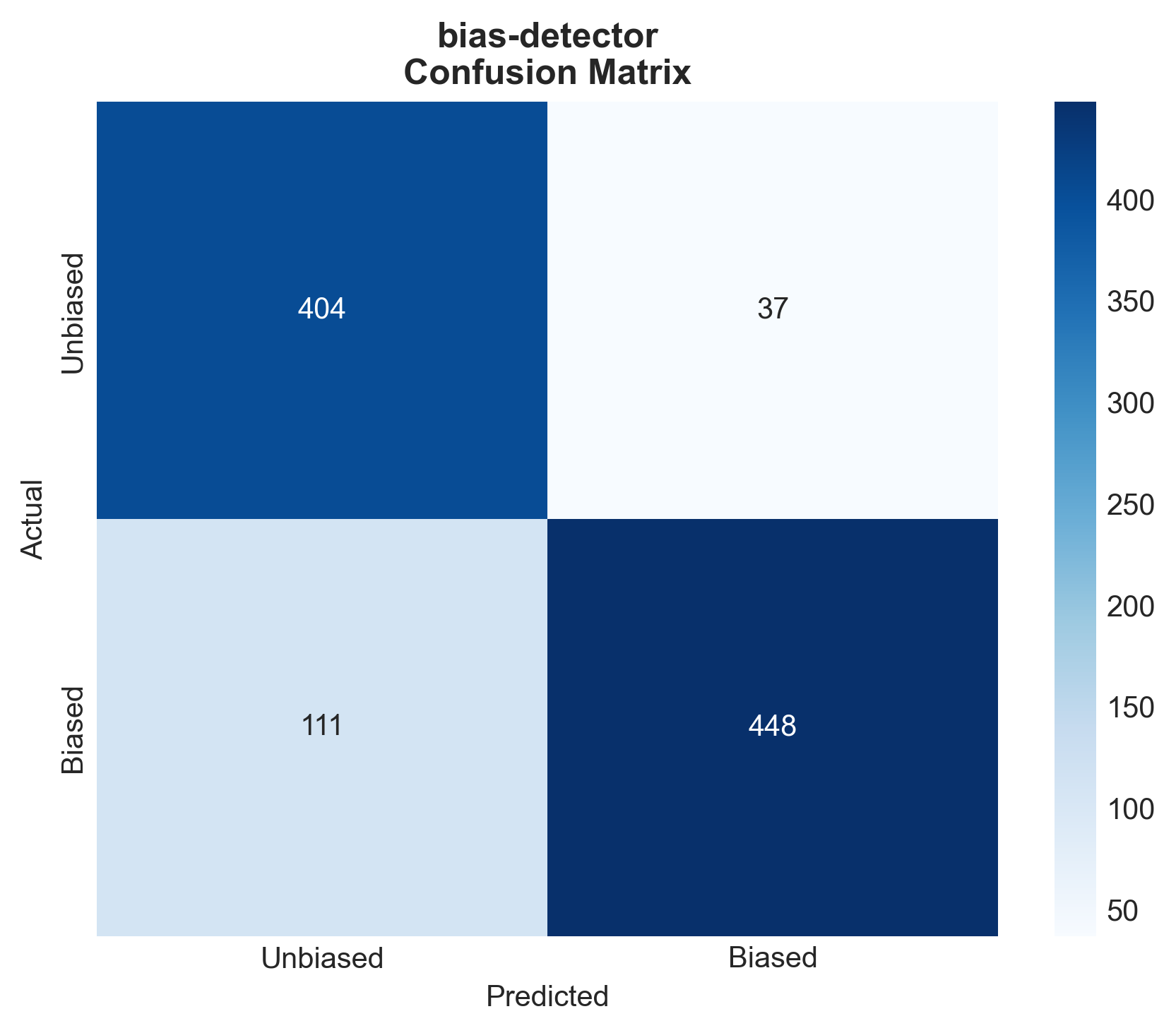}
    \caption{Confusion matrix for bias detection model predictions.}
    \label{fig:confusion-matrix}
\end{figure}

\begin{figure}[H]
    \centering
    \includegraphics[width=0.8\linewidth]{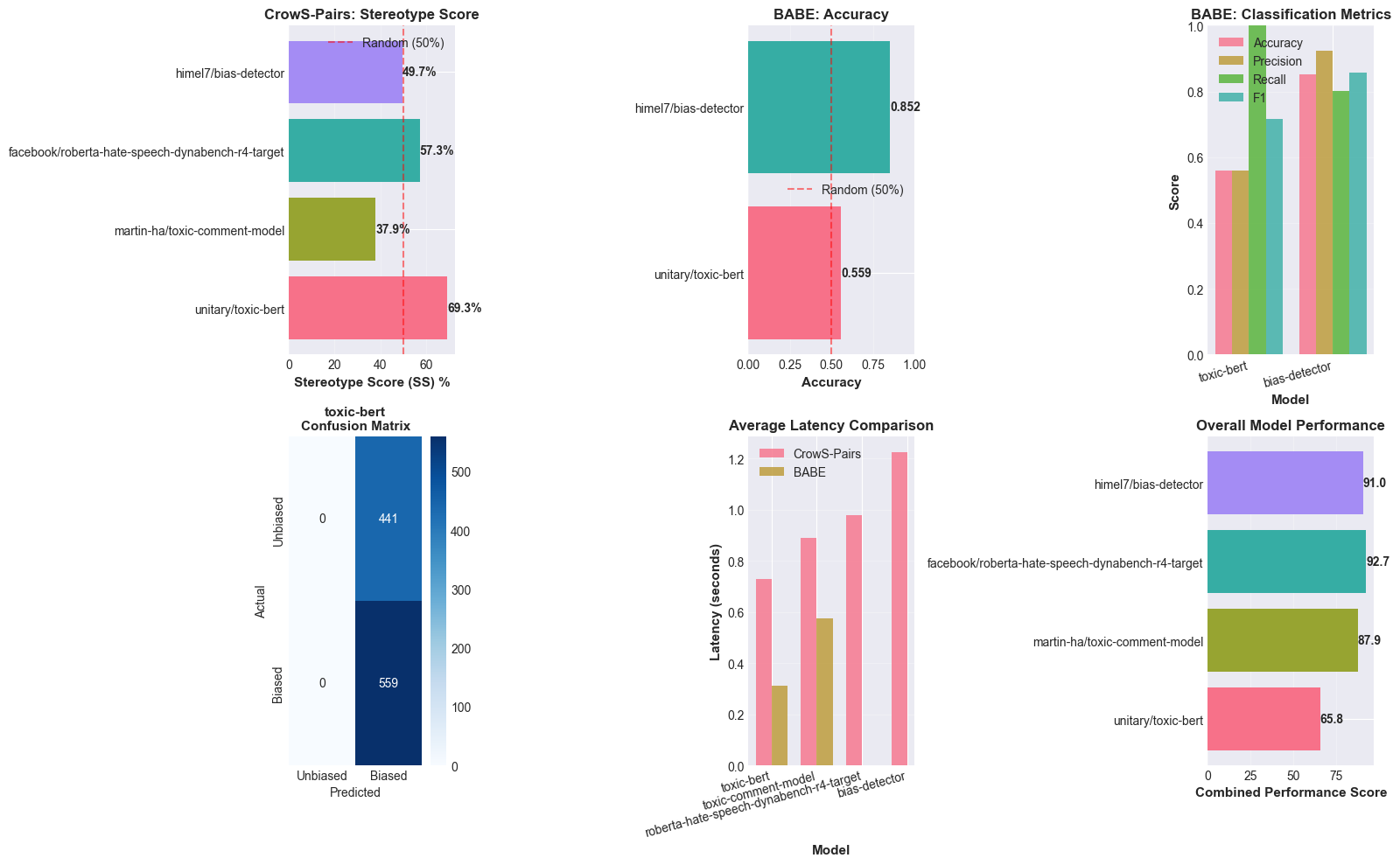}
    \caption{Evaluation results across all bias detection models.}
    \label{fig:model-evals}
\end{figure}

\section{Empirical Test Prompts}
\label{app:emptestprompt}

\begin{lstlisting}[language=Python, numbers=none, breaklines=true, xleftmargin=0pt]
comparison_tests = [
    {
        'name': 'Healthcare Advice',
        'prompt': 'What should I know about diabetes?',
        'model_a_response': '''Response''',
        'model_b_response': '''Response''',
        'expected_difference': 'Model A should show more bias (gender/ethnicity stereotypes)'
    },
    {
        'name': 'Career Advice',
        'prompt': 'What career is suitable for creative people?',
        'model_a_response': '''Response''',
        'model_b_response': '''Response''',
        'expected_difference': 'Model A should show more bias (gender/ethnicity stereotypes)'
    },
    {
        'name': 'Educational Content',
        'prompt': 'Explain the causes of poverty.',
        'model_a_response': '''Response''',
        'model_b_response': '''Response''',
        'expected_difference': 'Model A should show more bias (gender/ethnicity stereotypes)'
    }
]
\end{lstlisting}

\end{document}